\documentclass[letterpaper,10pt,conference]{ieeeconf}
\IEEEoverridecommandlockouts
\usepackage{algorithm}
\usepackage{algpseudocode}
\usepackage{amsmath}
\usepackage{amssymb}
\usepackage{bm}
\usepackage{caption}
\usepackage{color}
\usepackage{colortbl}
\usepackage{dsfont}
\usepackage{epsfig}
\usepackage{float}
\usepackage{graphics}
\usepackage{subcaption}
\usepackage{times}
\usepackage{url}

\newcommand{\commentout}[1]{}
\newcommand{\junk}[1]{}

\newcommand{\etal}{\emph{et al.}}

\newcommand{\bx}{{\bf x}}

\newcommand{\eps}{\varepsilon}

\newcommand{\realset}{\mathbb{R}}

\newcommand{\abs}[1]{\left|#1\right|}

\newcommand{\I}[1]{\mathds{1} \! \left\{#1\right\}}

\newcommand{\normw}[2]{\left\|#1\right\|_{#2}}

\newcommand{\set}[1]{\left\{#1\right\}}

\begin{document}

\title{\LARGE \bf Learning from a Single Labeled Face and a Stream of Unlabeled Data}

\author{\parbox{3in}{\centering
Branislav Kveton \\
Technicolor Labs \\
Palo Alto, CA \\
{\tt\small branislav.kveton@technicolor.com}}
\hspace*{0.5in}
\parbox{3in}{\centering
Michal Valko \\
Inria Lille - Nord Europe, team SequeL \\
Lille, France \\
{\tt\small michal.valko@inria.fr}}}

\maketitle
\thispagestyle{empty}
\pagestyle{empty}

\begin{abstract}
Face recognition from a single image per person is a challenging problem because the training sample is extremely small. We study a variation of this problem. In our setting, only a single image of a single person is labeled, and all other people are unlabeled. This setting is very common in authentication on personal computers and mobile devices, and poses an additional challenge because it lacks negative examples. We formalize our problem as one-class classification, and propose and analyze an algorithm that learns a non-parametric model of the face from a single labeled image and a stream of unlabeled data. In many domains, for instance when a person interacts with a computer with a camera, unlabeled data are abundant and easy to utilize. We show how unlabeled data can help in learning better models and evaluate our method on 43 people. The people are identified 90\% of the time at nearly zero false positives. This is 15\% more often than by Fisherfaces at the same false positive rate. Finally, we conduct a comprehensive sensitivity analysis of our method and provide a guideline for setting its parameters.
\end{abstract}

\section{Introduction}
\label{sec:introduction}

Face recognition from a single image per person is a hard problem because the training sample is extremely small \cite{tan06face}. Yet this setting is very common in practice and therefore has been of great interest. For instance, extensive databases with one labeled image per person already exist, such as those for ID cards, and face recognition from these data could enable population-wide security screening at airports. The challenge in learning from a single labeled image is that the appearance of the face changes due to many factors, such as aging, facial expressions, or growing a mustache. In general, such changes are hard to model, especially from a single image per person.

Face recognition research has made many advances due to learning discriminative projections \cite{zhao03face} and all state-of-the-art methods employ them in one way or another. Unfortunately, learning of high-quality discriminative projections requires a lot of labeled data. Therefore, it is not surprising that state-of-the-art face recognition methods perform poorly when only a single image per person is labeled (Section~\ref{sec:single image face recognition}). This problem becomes even more challenging when only a single \mbox{person is} labeled, and all other people are unlabeled. This is the setting considered in our paper.

We study face recognition from a single labeled image per person in the online setting. In addition to the labeled image, we observe a stream of unlabeled data, for instance recorded by a video camera. In this setting, the lack of labeled images can be compensated for by a large amount of unlabeled data. Computer vision problems usually exhibit a low-dimensional manifold structure \cite{he05face} and these data can be used to learn it. We propose a new face recognition method, \emph{online manifold tracking (OMT)}, that learns the structure of the manifold on-the-fly and can adapt to changes in data. The time and space complexity of our approach are bounded and do not increase with time. We compare our approach to several baselines and demonstrate its superiority. Finally, we evaluate its sensitivity to the setting of the parameters and discuss how to set them.

Online manifold tracking has several advantages. First, the algorithm is relatively easy to implement. Second, it does not require extensive offline training and is sufficiently fast to run in real time. In Sections~\ref{sec:experiments generalization radius} and \ref{sec:experiments representative faces}, we show that OMT recognizes faces in as little as 0.05 second on average. Third, our approach is non-parametric. We make no assumptions on recognized faces, and can adapt to various facial expressions and poses. Finally, our method is by design robust to outliers and thus suitable for open-world domains.

Non-parametric learning tends to be viewed as an alternative to learning with sophisticated features. We want to stress that our approach is complementary and benefits from better features (Section~\ref{sec:quality}). Similarly, we believe that most face recognition algorithms, which rely on discriminative features, could benefit from adaptation and handling the concept drift. This work shows how to incorporate such features into these algorithms.

\section{Face recognition from a single labeled face}
\label{sec:single image face recognition}

Face recognition from a \emph{single image per person} is a difficult problem \cite{tan06face} because all state-of-the-art face recognizers rely on a large number of training data, which are unavailable in this setting. In Fisherfaces \cite{belhumeur97eigenfaces}, two or more images of the person are needed to estimate the within-class variance. This problem is ill posed when only one training face is available. In Laplacianfaces \cite{he05face}, several images of the same person are necessary to estimate the low-dimensional manifold of faces. When only one image per person is available, Laplacianfaces reduce to maximizing the between-class variance \cite{tan06face} and are similar to eigenfaces. Eigenfaces \cite{turk91face} are maximum variance projections of data obtained by principal component analysis (PCA).

In this work, we study a variation of face recognition from a single image. In our setting, only one image of one person is labeled, and many other people are unlabeled. This setting is common in open-world domains, where the class of other people is hard to model explicitly. For instance, in face-based authentication on a computer, the owner of the computer has to be modeled but it is hard, even impossible, to individually model all other people. A major challenge in this problem is the lack of negative examples. Therefore, the problem cannot be directly formulated as learning a discriminator of positive and negative examples, as is common in face recognition \cite{zhao03face}.

\emph{One-class classification} \cite{tax01thesis} is a natural way of formulating our problem. In one-class classification, the goal is to learn a hypersphere that covers positive examples. Nearest-neighbor (NN) classification with one positive example is the simplest instance of such techniques. This classifier can be written as:
\begin{align}
  f^{\mathrm{nn}}_R(\bx) = \left\{
  \begin{array}{l l}
    1 & d(\bx, \bx_l) \leq R \\
    0 & \mathrm{otherwise,}
  \end{array}
  \right.
  \label{eq:NN classifier}
\end{align}
where $\bx_l$ is the labeled example, $d(\cdot, \cdot)$ is a distance function, and $R$ is the radius of the hypersphere. In this work, we refer to $R$ as a \emph{generalization radius} and assume that the distance $d(\cdot, \cdot)$ is Euclidean.

The accuracy of one-class classifiers is typically measured by the \emph{true positive (TPR)} and \emph{false positive (FPR) rates}. The TPR is the fraction of positives classified as positives and the FPR is the fraction of negatives classified as positives. In the NN classifier $f^{\mathrm{nn}}_R(\bx)$, both rates monotonically increase with the generalization radius $R$. The radius $R$ should be set such that the classifier has high TPR and acceptably low FPR.

\section{Face recognition from a stream of unlabeled faces}
\label{sec:unlabeled stream face recognition}

Many face recognition algorithms can be viewed as batch-mode NN classifiers in some metric space $d(\cdot, \cdot)$ (Section~\ref{sec:existing work}). This space is defined by discriminative features. In principle, it is hard to learn good discriminative features when only one example is labeled (Section~\ref{sec:single image face recognition}). So instead, we take advantage of the structure of unlabeled data and learn which part of the feature space belongs to the same person as the labeled face $\bx_l$.

In particular, we learn a non-parametric predictor of a face from a single labeled face and a stream of unlabeled images. This problem is challenging for a few reasons. First, the data are unlabeled and may contain images of other people. As a result, it is necessary to be cautious when generalizing. This is why state-of-the-art face recognizers often perform poorly in practice. Second, the sequence of unlabeled faces may be long, and even infinite. Therefore, our non-parametric model should be compact and sublinear, or constant, in the number of observed faces.

Formally, our learning problem is modeled as a repeating game against a potentially adversarial nature. At each step $t$ of this game, we observe an example $\bx_t$ and then predict its label based on all observations $\bx_1, \dots, \bx_t$ up to time $t$. This problem is challenging because only one example is labeled. Therefore, if we want to learn in this setting, we have to rely on \emph{indirect} forms of \emph{feedback}, such as the similarity between the observations $\bx_1, \dots, \bx_t$.

This section is organized as follows. In Section~\ref{sec:OMT}, we show how to compactly represent a potentially infinite stream of data. In Sections~\ref{sec:inference} and \ref{sec:algorithm}, we discuss how to infer the identity of a person based on our compact representation. In Section~\ref{sec:parameterization}, we discuss how to set the parameters of our algorithm.

\subsection{Online manifold tracking}
\label{sec:OMT}

\begin{algorithm}[t]
  \caption{Online manifold tracking.}
  \label{alg:OMT}
  \begin{algorithmic}
    \State {\bf Input:}
    \State \quad Representative faces $u_t$
    \State \quad Observed face $\bx_t$
    \State \quad Generalization radius $R$
    \State \quad Cover radius $r$
    \State
    \If {$(d(\bx_t, \bx_l) \leq R)$}
      \If {$(\forall i \in u_t \ (d(\bx_t, \bx_i) > r))$}
        \State $u_{t + 1} \gets u_t \cup \set{t}$
      \Else
        \State $u_{t + 1} \gets u_t$
      \EndIf
      \While{$(\abs{u_{t + 1}} = k + 1)$}
        \State $r \gets 2 r$
        \State $u_\mathrm{old} \gets u_{t + 1}$
        \State Greedily select face indices $u_{t + 1} \subseteq u_\mathrm{old}$ such that:
        \State \qquad $\forall i \in u_\mathrm{old} \ \exists j \in u_{t + 1} \ (d(\bx_i, \bx_j) \leq r)$
        \State \qquad $\forall i \in u_{t + 1} \ \forall j \in (u_{t + 1} \setminus i) \ (d(\bx_i, \bx_j) > r)$
      \EndWhile
    \EndIf
    \State
    \State {\bf Output:}
    \State \quad Representative faces $u_{t + 1}$
    \State \quad Cover radius $r$
  \end{algorithmic}
\end{algorithm}

One way of summarizing data is by mapping each example to the closest representative example. This approach is known as \emph{data quantization} \cite{gray98quantization} and the representative examples can be found by various techniques, such $k$-means clustering and random sampling. In our setting, we want to summarize data on-the-fly. Two popular methods for online data quantization are online $k$-center clustering \cite{charikar97incremental} and cover trees \cite{beygelzimer06cover}.

In this paper, we quantize faces by online $k$-center clustering \cite{charikar97incremental}. At time $t$, all previously seen faces are summarized by indices $u_t$ of up to $k$ \emph{representative faces}. The indices are updated as follows. If the face $\bx_t$ at time $t$ is at least $r$ away from all representative faces $u_t$, $u_{t + 1} = u_t \cup \set{t}$. Otherwise, $u_{t + 1} = u_t$. Finally, when $\abs{u_{t + 1}} = k + 1$, the \emph{cover radius} $r$ is doubled and the representative faces are repartitioned such that no two faces are closer than $r$.

Our implementation of online $k$-center clustering is shown in Algorithm~\ref{alg:OMT}. Note that the example $\bx_t$ is quantized only if it is sufficiently close to the labeled example $\bx_l$, $d(\bx_t, \bx_l) \leq R$. Therefore, the \emph{generalization radius} $R$ essentially controls how much space is covered. In practice, it should be set such that we do not cover parts of the space that are too far away from the labeled example $\bx_l$ and may be irrelevant when we extrapolate from it. More discussion on how to set the value of $R$ can be found in Section~\ref{sec:parameterization}.

Because online $k$-center clustering provides guarantees on the error of its approximation \cite{charikar97incremental}, we can bound the error of Algorithm~\ref{alg:OMT}. In particular, at any time $t$, the distance between any previously seen face and the closest representative face:
\begin{align}
  d_{\max}^t(u_t) = \max_{\stackrel{\scriptstyle i < t}{d(\bx_i, \bx_l) \leq R}}
  \min_{j \in u_t} d(\bx_i, \bx_j)
  \label{eq:cover error}
\end{align}
is bounded by $2 r$. The \emph{error of the cover} $d_{\max}^t(u_t)$ is always smaller than 8 times of that of the optimal cover. The optimal cover of cardinality $k$ minimizes $d_{\max}^t(\cdot)$ and its computation is NP hard.

Because the error $d_{\max}^t(u_t)$ is bounded, we can also bound the error of our identify inference algorithm in Section~\ref{sec:inference}. This proof would proceed along the lines of Valko \etal~\cite{valko10online}.

\subsection{Inference}
\label{sec:inference}

\begin{figure}[t]
  \centering
  \includegraphics[width=2.9in, bb=0.5in 0.5in 3.7in 3.7in]{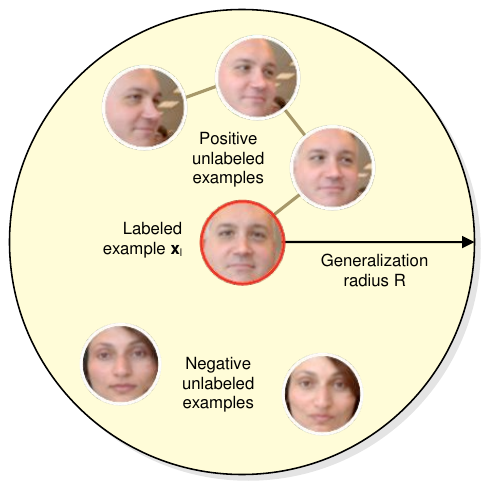}
  \vspace{0.1in}
  \caption{An illustration of the face manifold tracked by OMT. The labeled example $\bx_l$ is shown in the middle.}
  \label{fig:inference}
\end{figure}

Identity inference on a manifold of faces can be formulated as a random walk on a graph, where the vertices are the faces and the edges are weighted by the similarity $w_{ij}$ of the faces \cite{balcan05application}. This random walk starts at an unlabeled face $\bx_i$, jumps to neighboring faces $\bx_j$ proportionally to their similarity $w_{ij}$, and is absorbed at labeled faces. The absorption probabilities $F \in [0, 1]^{\abs{u} \times \abs{l}}$ can be computed as:
\begin{align}
  F = (L_{uu})^{-1} W_{ul},
  \label{eq:HS}
\end{align}
where $W \in \realset^{n \times n}$ is the matrix of pairwise face similarities, $L$ is its combinatorial Laplacian, $l$ is the set of labeled faces, $u$ is the set of unlabeled faces, and $n$ is the number of faces. Equation~\ref{eq:HS} is well known as the \emph{harmonic solution (HS)} and is a basis for many semi-supervised learning algorithms \cite{zhu03semisupervised}.

The main challenge in computing the harmonic solution in our setting is that we have only one labeled positive example $\bx_l$ (Section~\ref{sec:single image face recognition}). Therefore, all random walks on the graph $W$ ultimately terminate in this example and the HS $F = \mathbf{1}_{\abs{u} \times 1}$ is meaningless. Note that our result is mathematically correct and is due to not modeling any other identity than that \mbox{of $\bx_l$.}

\begin{algorithm}[t]
  \caption{Identity inference.}
  \label{alg:inference}
  \begin{algorithmic}
    \State {\bf Input:}
    \State \quad Representative faces $u_{t + 1}$
    \State \quad Observed face $\bx_t$
    \State \quad Generalization radius $R$
    \State \quad Recognition threshold $\eps$
    \State
    \If {$(d(\bx_t, \bx_l) \leq R)$}
      \State $u \gets u_{t + 1}$
      \State $v \gets u \cup \set{l}$
      \State $W \gets \abs{v} \times \abs{v}$ similarity matrix of faces $v$
      \State $D \gets \abs{v} \times \abs{v}$ diagonal matrix s.t. $D_{ii} = \sum_j W_{ij}$
      \State $L \gets D - W$
      \State Compute the probability that the faces $v$ are positives:
      \State \qquad ${\bf f} \gets (L_{uu} + \gamma I_u)^{-1} W_{ul}$
      \State $j \gets \arg\min_{i \in u} d(\bx_t, \bx_i)$
      \State $\hat{y}_t \gets \I{f_j > \eps}$
    \Else
      \State $\hat{y}_t \gets 0$
    \EndIf
    \State
    \State {\bf Output:}
    \State \quad Identity $\hat{y}_t$ of the face $\bx_t$
  \end{algorithmic}
\end{algorithm}

We do not want to explicitly represent negative examples. This is because we want to model how the person looks and do not want to waste our resources on modeling other people. However, we need to introduce some notion of dissimilarity. For instance, if the vertex $\bx_l$ cannot be reached from another vertex $\bx_t$ in a small number of random jumps, these vertices may not be similar. To achieve this behavior, we introduce a special \emph{sink} vertex $\bx_0$. This vertex absorbs all random walks that reach it and is connected to all unlabeled vertices $i \in u$ by weighted edges $w_{i0} = \gamma$, where $\gamma$ is a tunable parameter. Therefore, not all random walks get absorbed by the labeled vertex $\bx_l$. The probability of being absorbed depends on the structure of $W$, $\gamma$, and the starting point of the random walk. Similarly to the HS (Equation~\ref{eq:HS}), the absorption probabilities ${\bf f} \in [0, 1]^{\abs{u} \times 1}$ can be computed in a closed form \cite{kveton10semisupervised}:
\begin{align}
  {\bf f} = (L_{uu} + \gamma I_u)^{-1} W_{ul},
  \label{eq:HS sink}
\end{align}
where $I_u$ is a $\abs{u} \times \abs{u}$ identity matrix. Our identity inference method is outlined in Algorithm~\ref{alg:inference}.

\subsection{Algorithm}
\label{sec:algorithm}

Our solution is an online learning algorithm. At time $t$, we quantize the face $\bx_t$ (Algorithm~\ref{alg:OMT}) and then infer its identity (Algorithm~\ref{alg:inference}). We refer to our technique as \emph{online manifold tracking (OMT)} because it approximately tracks the manifold of faces and then utilizes it to build a better face recognizer. An illustration of the tracked manifold is shown in Figure~\ref{fig:inference}.

Each step of our algorithm consumes $O(k^3)$ time because online $k$-center clustering takes $O(k)$ time and the harmonic solution can be computed in $O(k^3)$ time, by solving $k$ linear equations. As a result, the time complexity of our algorithm is independent of time $t$. Note that when the similarity matrix $W$ is $O(k)$ sparse, the time complexity of computing the HS on $W$ is $O(k^2)$. In addition, many fast approximate solutions exist.

\subsection{Parameterization}
\label{sec:parameterization}

Our method has several tunable parameters. In this section, we discuss how to set these parameters and explain how they affect the behavior of our algorithm. In Section~\ref{sec:experiments}, we show that many of these parameters do not have to be set perfectly to achieve good performance.

The generalization radius $R$ controls extrapolation to unlabeled data. Larger values of $R$ result in farther extrapolation. Technically, the radius $R$ determines the maximum TPR and FPR of our method. Note that these are the same as the TPR and FPR of the classifier $f^{\mathrm{nn}}_R(\bx)$ (Equation~\ref{eq:NN classifier}) with the same radius $R$. In practice, $R$ should be set to the minimum value such that the maximum TPR and FPR are high and relatively low, respectively. In Section~\ref{sec:experiments generalization radius}, we conduct an experiment that shows how the generalization radius $R$ impacts learning.

The maximum number of representative faces $k$ trades off the error of the cover (Equation~\ref{eq:cover error}) for the computational cost of inference. In general, as $k$ increases, the error of the cover decreases and the time complexity of our approach increases, cubically with $k$. In Section~\ref{sec:experiments representative faces}, we conduct an experiment that illustrates these trends.

The main parameter that controls the TPR and FPR of our algorithm is the recognition threshold $\eps$ (Algorithm~\ref{alg:inference}). Both the TPR and FPR increase as $\eps$ decreases. So the ROC curve for our method can be generated by varying $\eps$. We adopt this methodology in the experimental section.

Graph-based inference algorithms \cite{zhu03semisupervised} tend to be sensitive to the choice of the graph and our method is not an exception. In our domain, the \emph{similarity} of faces $\bx_i$ and $\bx_j$ is computed as:
\begin{align}
  w_{ij} = \exp \left[- d^2(\bx_i, \bx_j) / (2 \sigma^2)\right],
  \label{eq:similarity}
\end{align}
where $\sigma$ is the \emph{heat parameter} and $d(\bx_i, \bx_j)$ is the distance of the faces. The \emph{distance} is defined as $d(\bx_i, \bx_j) = \normw{\bx_i - \bx_j}{2}$, where $\bx_i$ and $\bx_j$ denote pixel intensities in $96 \times 96$ images of faces. The intensities are rescaled such that $\max_{\bx} \normw{\bx}{2} = 1$. So the maximum distance between any two faces is two. The distance between consecutive faces in our datasets is usually less than 0.1. We set the heat parameter $\sigma$ to 0.03 and so this distance is about $3 \sigma$. Our setting is motivated by a statistical rule that events that are $3 \sigma$ away from the mean are unlikely. We experimented with other values $\sigma$, both 0.025 and 0.035, and all trends in our experiments remained the same.

The similarity to the sink is $\gamma = \exp[- 3^2 / 2]$. This setting can be interpreted as follows. When two faces are closer than $3 \sigma$, the probability that a random transition between the faces terminates in the sink $\bx_0$ is less than:
\begin{align}
  \frac{\gamma}{\gamma + w_{ij}} \leq
  \frac{\gamma}{\gamma + \exp[- 3^2 / 2]} = \frac{1}{2}.
  \label{eq:gamma 3}
\end{align}
On the other hand, when two faces are more than $4 \sigma$ and $5 \sigma$ away, the probability of terminating in the sink $\bx_0$ is \mbox{at least:}
\begin{align}
  \frac{\gamma}{\gamma + w_{ij}} \geq
  \frac{\gamma}{\gamma + \exp[- 4^2 / 2]} \approx 0.9707
  \label{eq:gamma 4}
\end{align}
and:
\begin{align}
  \frac{\gamma}{\gamma + w_{ij}} \geq
  \frac{\gamma}{\gamma + \exp[- 5^2 / 2]} \approx 0.9997,
  \label{eq:gamma 5}
\end{align}
respectively. In other words, faces that are more distant than $3 \sigma$ are likely to be perceived as different, and the probability of being different increases exponentially with their distance.

\section{Experiments}
\label{sec:experiments}

We evaluate our method on video recordings of 43 people (Section~\ref{sec:dataset}). Our experimental results support two claims. First, we show that online learning from a single labeled face and unlabeled faces performs better than supervised learning (Section~\ref{sec:quality}). Second, we demonstrate that our approach is complementary to learning with better features. In particular, we show that OMT with Fisherfaces outperforms both OMT and Fisherfaces when used separately. Finally, we conduct a comprehensive sensitivity analysis of our method and discuss how to parameterize it (Sections~\ref{sec:experiments generalization radius} and \ref{sec:experiments representative faces}). We observe that OMT performs robustly even when its parameters are not set optimally.

\begin{figure*}[t]
  \centering
  \begin{subfigure}[b]{6.8in}
    \centering
    \includegraphics[width=6.8in, viewport=0in 4in 8.5in 7in]{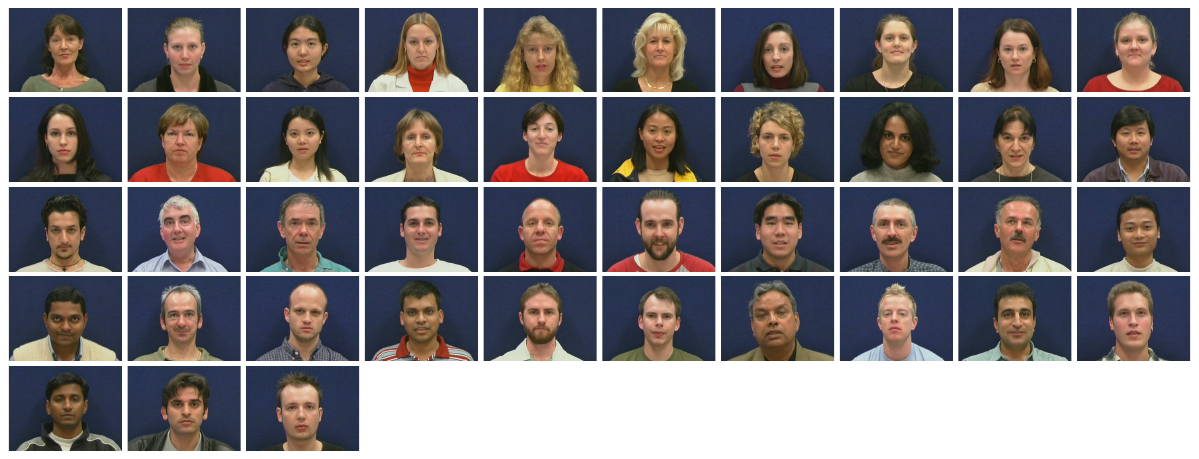}
    \caption{People in the dataset.}
    \label{fig:faces}
    \vspace{0.1in}
  \end{subfigure}
  \begin{subfigure}[b]{6.8in}
    \centering
    \includegraphics[width=6.8in, viewport=0in 4.75in 8.5in 6.25in]{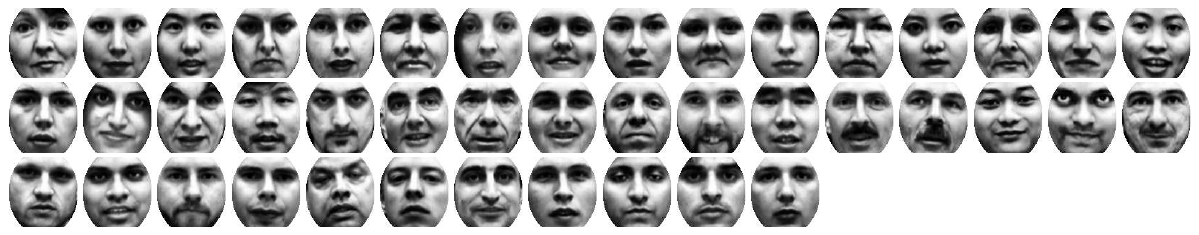}
    \caption{One labeled face per person.}
    \label{fig:labeled faces}
    \vspace{0.1in}
  \end{subfigure}
  \begin{subfigure}[b]{6.8in}
    \centering
    \includegraphics[width=6.8in, viewport=0in 5in 8.5in 6in]{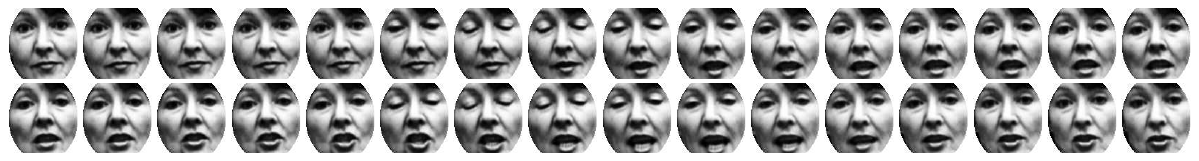}
    \caption{A sequence of faces in one video.}
    \label{fig:face stream}
    \vspace{0.1in}
  \end{subfigure}
  \begin{subfigure}[b]{6.8in}
    \centering
    \includegraphics[width=6.8in, viewport=0in 4.5in 8.5in 6.5in]{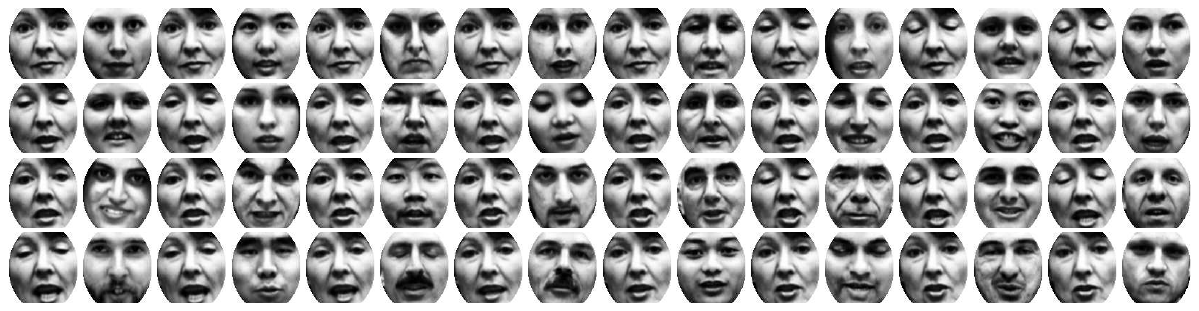}
    \caption{A noisy sequence of faces. The odd faces belong to the original video (Figure~\ref{fig:face stream}) and the even ones are chosen randomly \mbox{from the} videos of the remaining 42 people.}
    \label{fig:noisy face stream}
    \vspace{0.1in}
  \end{subfigure}
  \caption{Images and faces in the VidTIMIT dataset.}
  \label{fig:dataset}
\end{figure*}

\subsection{Dataset}
\label{sec:dataset}

The VidTIMIT dataset \cite{sanderson09multiregion} is comprised of video and the corresponding audio recordings of 43 people (Figure~\ref{fig:faces}) that recite short sentences. The dataset was recorded in 3 sessions. The delay between Sessions~1 and 2 is 7 days, and the delay between Sessions~2 and 3 is 6 days. Each person is asked to recite ten sentences: 6 in Session~1, 2 in Session~2, and 2 in Session~3. The recording was done in an office environment using a broadcast quality digital video camera. The video of each person is a sequence of 512 x 384 images. The average length of the video is 1062 images. The primary variations in our data are in facial expressions and time, since the dataset is comprised of three separate recordings.

Faces in the images are detected by OpenCV \cite{opencv-library}, turned into grayscale, resized to $96 \times 96$ pixels, cropped, and finally we equalize their histograms. We label one image per person (Figure~\ref{fig:labeled faces}).

\subsection{Methodology}
\label{sec:methodology}

\begin{figure*}[t]
  \centering
  \includegraphics[width=6.8in, viewport=0in 4.5in 8.5in 6.5in]{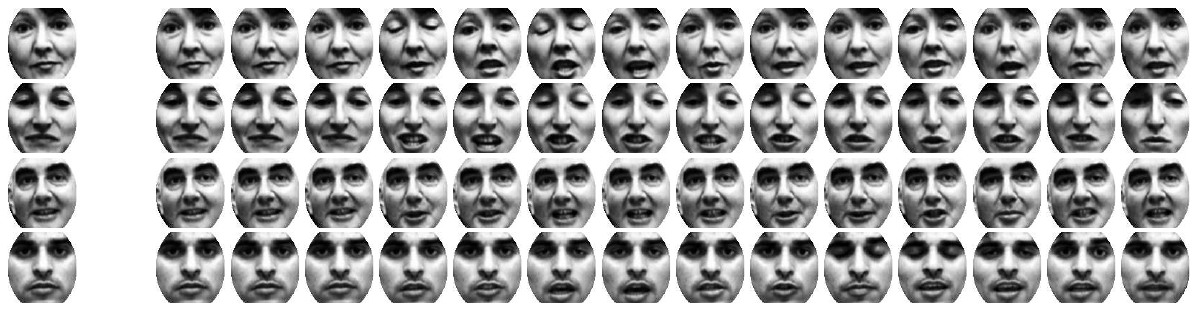}
  \vspace{0.1in}
  \caption{Representative faces learned by OMT for Person 1, 15, 22, and 42. The four leftmost faces are the labeled examples.}
  \label{fig:representative faces}
\end{figure*}

All experiments are conducted on 43 video traces from the VidTIMIT dataset (Section~\ref{sec:dataset}). In each video, one person recites 10 sentences and no other person appears. This setting does not seem challenging because the identity in each video frame can be predicted by tracking the face from the labeled image. To make the videos more realistic, we add outliers to them. In particular, after each frame in the video (Figure~\ref{fig:face stream}), we insert a randomly selected image from the remaining 42 videos (Figure~\ref{fig:noisy face stream}). The new videos are challenging because a half of the frames are negatives, people that do not belong to the modeled class. Moreover, two consecutive faces never belong to the same person, and therefore face recognition by tracking would perform poorly in this setting. However, note that the videos are still temporarily smooth in the sense that two consecutive positives are similar. OMT can identify this pattern and learns from it.

The quality of solutions is measured by their TPR and FPR at various operating points on the ROC curve. The operating points of the NN classifier are obtained by varying the radius $R$ (Equation~\ref{eq:NN classifier}). The operating points of OMT are computed by varying the recognition threshold $\eps$ (Algorithm~\ref{alg:inference}). In each video, we compute the TPR and FPR, and then average them over all videos. The generalization radius $R$ and the number of representative faces $k$ in OMT are by default 0.3 and 300, respectively. The sensitivity to the setting of these parameters is studied in Sections~\ref{sec:experiments generalization radius} and \ref{sec:experiments representative faces}.

\subsection{Quality of solutions}
\label{sec:quality}

\begin{figure}[t]
  \centering
  \includegraphics[width=3.2in, viewport=2.25in 4in 6.25in 7in]{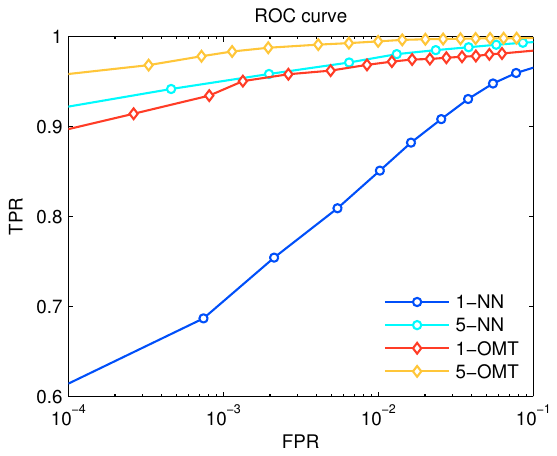}
  \caption{Comparison of the NN and OMT recognizers that are trained from 1 and 5 labeled faces.}
  \label{fig:baseline labels}
\end{figure}

\begin{figure}[t]
  \centering
  \includegraphics[width=3.2in, viewport=2.25in 4in 6.25in 7in]{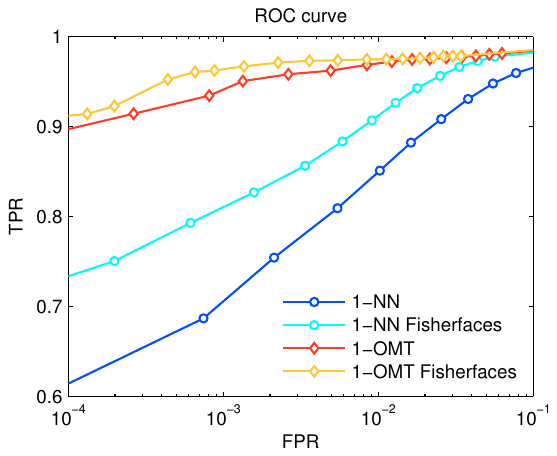}
  \caption{Comparison of the NN and OMT recognizers that are trained on pixel intensities and projections on 64 Fisherfaces.}
  \label{fig:baseline fisherfaces}
\end{figure}

In the first experiment, we compare our algorithm to three baselines. The first baseline is a 1-NN classifier (Equation~\ref{eq:NN classifier}) and we compare to it to illustrate the benefit of learning from unlabeled faces. The second baseline is a 5-NN classifier and it shows how much labeled data are needed to learn as good predictor as using our algorithm. The last baseline is a 1-NN classifier in the space of 64 Fisherfaces. The Fisherfaces are computed from 43 labeled faces (Figure~\ref{fig:labeled faces}), one per person. Note that the within-class scatter matrix in our problem is all zeros. Therefore, it cannot be used in Fisherfaces (Section~\ref{sec:single image face recognition}) and we substitute it for an identity matrix. We experimented with various numbers of Fisherfaces and 64 yields the largest area under the ROC curve in Figure~\ref{fig:baseline labels}. Our results are shown in Figures~\ref{fig:baseline labels} and \ref{fig:baseline fisherfaces}. We observe four major trends.

First, OMT learns a pretty accurate predictor. The TPR of OMT at $10^{-4}$ FPR is 0.89. In other words, OMT recognizes people most of the time at nearly zero false positives. A few examples of correctly identified faces are shown in Figure~\ref{fig:representative faces}. Many faces are quite different from the original labeled face. Each video is processed in 45 seconds on average. Therefore, an average face is recognized in $45 / (2 \cdot 1062) \approx 0.02$ second, essentially in real time.

Second, OMT performs significantly better than the 1-NN baseline. The TPR of OMT at $10^{-4}$ FPR is 0.89, 50\% higher than that of the baseline. Note that both OMT and the 1-NN classifier are trained using the same amount of labeled data. So our comparison demonstrates the benefit of learning from unlabeled data. Finally, we plot the ROC curve for the 5-NN classifier (Figure~\ref{fig:baseline labels}) and note that it is similar to that of OMT. As a result, we may conclude that OMT learns the equivalent of 5 labeled faces.

Third, OMT performs much better than the 1-NN baseline on Fisherfaces. At low FPRs, the improvement in the TPR is in double digits. In contrast, note that most holistic methods outperform Fisherfaces and eigenfaces only in the low single digits \cite{tan06face}.

Fourth, OMT improves with more labeled faces and better features, similarly to other face recognition algorithms. For instance, Figure~\ref{fig:baseline labels} shows that the NN baseline improves a lot when the number of labeled faces increases from 1 to 5. The TPR of OMT, which is already in the low nineties, increases in this case by about 5\%, and is higher than the new baseline at all FPRs. Figure~\ref{fig:baseline fisherfaces} shows that the 1-NN baseline improves when the original feature space is substituted for Fisherfaces. The TPR of OMT increases in this case by 2\% at low FPRs, and is higher than the new baseline at all FPRs.

\subsection{Generalization radius $R$}
\label{sec:experiments generalization radius}

\begin{figure}[t]
  \centering
  \includegraphics[width=3.2in, viewport=2.25in 3in 6.25in 8in]{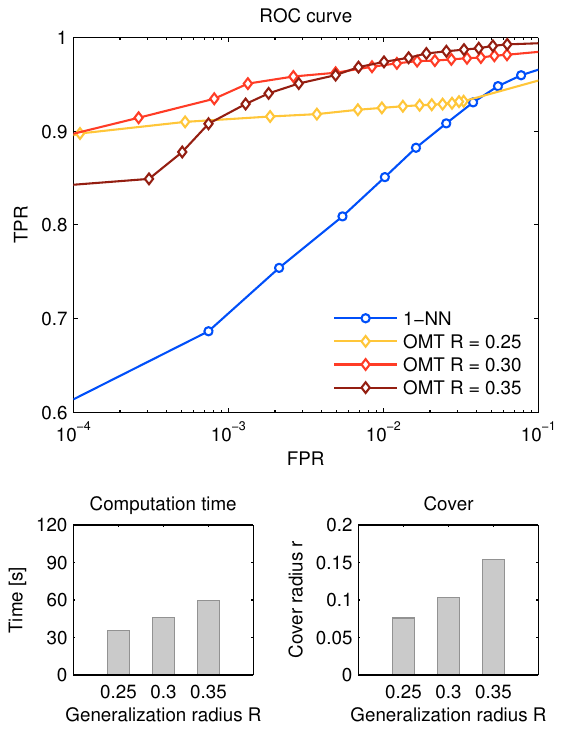}
  \caption{Varying the generalization radius $R$ in OMT. For each value $R$, we report the ROC curve, the computation time, and the cover radius $r$.}
  \label{fig:var generalization radius}
\end{figure}

In the second experiment, we study how the generalization radius $R$ affects the behavior of OMT. Our results are shown in Figure~\ref{fig:var generalization radius}. We observe several trends.

At all FPRs, the TPR for $R = 0.3$ is higher than the TPR for $R = 0.25$. This trend can be explained as follows. About 8\% of positives are farther from the labeled example $\bx_l$ than 0.25. Because the TPR for $R = 0.3$ is always higher than the TPR for $R = 0.25$, many of these positives can be classified correctly at nearly zero false positives. So the generalization radius of $R = 0.25$ is too restrictive.

At low FPRs, the TPR for $R = 0.3$ is higher than the TPR for $R = 0.35$. This trend can be explained as follows. Barely 2\% of positives are farther from the labeled example $\bx_l$ than 0.3. As a result, the potential increase in true positives when the radius $R$ increases beyond 0.3 is small, and in our results it is outweighed by the increase in false negatives. Therefore, $R = 0.3$ yields a higher TPR at low FPRs. Nevertheless, we note that OMT performs acceptably well for all tested values of $R$.

Finally, note that as the generalization radius $R$ increases, the cover radius $r$ and computation time increase. The radius $r$ increases since the covered space, $\bx_t$ such that $d(\bx_t, \bx_l) \leq R$, increases but the number of faces $k$ that cover it remains constant. The computation time increases because more faces satisfy $d(\bx_t, \bx_l) \leq R$, and must be quantized and classified.

\subsection{Number of representative faces $k$}
\label{sec:experiments representative faces}

\begin{figure}[t]
  \centering
  \includegraphics[width=3.2in, viewport=2.25in 3in 6.25in 8in]{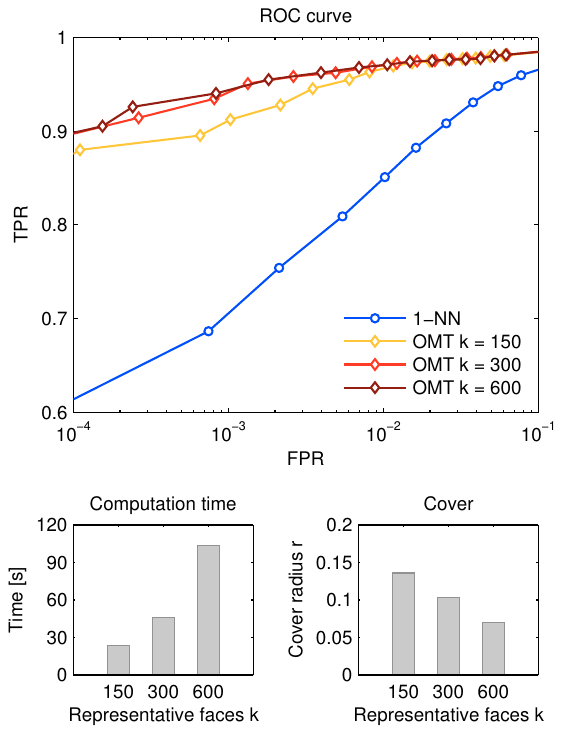}
  \caption{Varying the number of representative faces $k$ in OMT. For each value $k$, we report the ROC curve, the computation time, and the cover radius $r$.}
  \label{fig:var representative faces}
\end{figure}

In the last experiment, we study how the behavior of OMT changes based on the number of representative faces $k$. Our results are reported in Figure~\ref{fig:var representative faces}. We observe several trends.

As the number of representative faces $k$ increases, both the accuracy of inference and computation time increase, and the cover radius $r$ decreases. The radius $r$ decreases because the covered space, $\bx_t$ such that $d(\bx_t, \bx_l) \leq R$, remains the same but the number of faces $k$ that cover it increases. As a result, the accuracy also increases. The computation time grows less than quadratically with $k$, significantly slower than suggested by the analysis of our method (Section~\ref{sec:algorithm}). The reason for this trend is that our feature vectors $\bx_t$ are long, $96^2$ entries, and their quantization dominates the computational cost. The amortized per-step cost of online $k$-center clustering is $O(k)$, which is in line with the observed trend.

We recommend that the number of representative faces $k$ be chosen as high as the computational resources allow. The more variable the face and environment, the larger the value of $k$. Finally, note that as few as 150 representative faces are sufficient to learn interesting patterns.

\section{Related work}
\label{sec:existing work}

In this section, we review related work on face recognition and online semi-supervised learning.

\subsection{Face recognition}
\label{sec:existing face recognition}

The state-of-the-art in face recognition \cite{zhao03face} advanced so far that face recognition is available in consumer products. Face recognition from a single image per person is still considered to be a hard problem and we study a variation of this problem \cite{tan06face}. According to Tan \etal~\cite{tan06face}, we propose a \emph{holistic method} because we identify faces based on the whole image, and do not extract local features. Existing holistic methods \cite{tan06face} either employ some form of PCA to extract features \cite{wu02face} or enlarge the training set, for instance by novel views of the face \cite{byemer95face}. The novel views are generated by transformations, which are learned from a separate training set that comprises all views.

We take a very different approach in this paper. This is the first work that shows how to learn computationally efficiently a non-parametric model of a face from a stream of unlabeled data and a single labeled face. Our method can be viewed as learning novel views of the face from unlabeled data. Unlike Beymer and Poggio \cite{byemer95face}, the method does not have an offline training phase and can learn concepts that are hard to model, such as aging or growing a mustache. The main disadvantage of our method is that it is data driven. Therefore, it may need a large amount of unlabeled data to learn. Such data may not be available in all domains.

Note that our approach is complementary to learning from more sophisticated features. In Section~\ref{sec:quality}, we apply OMT to Fisherfaces and demonstrate that the new approach yields better results than each method separately.

\subsection{Online semi-supervised learning}
\label{sec:existing OSSL}

In machine learning, online learning from partially labeled data is known as \emph{online semi-supervised learning}. This problem has been formulated and solved in various ways, such as \noindent boosting, regularization of support vector machines (SVMs), and learning on graphs. \emph{Online semi-supervised boosting} \cite{grabner08semisupervised} is a variation of boosting, in which unlabeled data are labeled greedily using the data adjacency graph and then employed in the standard boosting fashion. \emph{Online manifold regularization of SVMs} \cite{goldberg08online} regularizes a max-margin classifier by the data adjacency graph. \emph{Online semi-supervised learning on a graph} \cite{valko10online} incrementally compresses the data adjacency graph and then infers labels of unlabeled examples based on this graph.

All of the above methods assume that at least two classes of examples are labeled and cannot be easily extended to our setting. Valko \etal~\cite{valko10online} and Balcan \etal~\cite{balcan05application} studied face recognition on similarity graphs from multiple labeled faces. This is the first work that studies face recognition on a graph from a single labeled image.

\section{Conclusions}
\label{sec:conclusions}

In this paper, we present online manifold tracking (OMT), a new online face recognition algorithm which is suitable for environments with minimal human supervision. In comparison to existing methods, which learn discriminative features, OMT relies on unlabeled data as the main form of feedback. We evaluate our method on a dataset of 43 people and show that it produces superior results. In addition, we demonstrate that OMT is complementary to learning with better features, such as Fisherfaces. Finally, we discuss how to parameterize our method and show that it is robust to a small perturbation of its parameters.

In this work, OMT is presented as a holistic method, where the whole face is treated as an input. In our future work, we plan to extend OMT to local facial features, such as the nose, eyes, and mouth. In the single-image-per-person setting, it is accepted that local methods outperform holistic methods \cite{tan06face}. We strongly believe that we can improve these methods even further by online adaptation, perhaps based on similarities in consecutive video frames.

\section{Acknowledgements}
\label{sec:acknowledgements}

This research work was supported by Ministry of Higher Education and Research, Nord-Pas de Calais Regional Council and FEDER through the ``contrat de projets {\'e}tat region 2007--2013", and by PASCAL2 European Network of Excellence. The research leading to these results has also received funding from the European Community's Seventh Framework Programme (FP7/2007-2013) under grant agreement n$^{\rm o}$ 270327 (project CompLACS).

\bibliographystyle{IEEEtran}
\bibliography{References}

@inproceedings{ turk91face,
  author = "Matthew Turk and Alex Pentland",
  title = "Face Recognition Using Eigenfaces",
  booktitle = "IEEE Conference on Computer Vision and Pattern Recognition",
  pages = "586-591",
  year = "1991"
}

@inproceedings{ byemer95face,
  author = "David Beymer and Tomaso Poggio",
  title = "Face Recognition from One Example View",
  booktitle = "Proceedings of the 5th International Conference on Computer Vision",
  pages = "500-507",
  year = "1995"
}

@article{ belhumeur97eigenfaces,
  author = "Peter Belhumeur and Jo{\~a}o Hespanha and David Kriegman",
  title = "Eigenfaces vs. {Fisherfaces}: Recognition Using Class Specific Linear Projection",
  journal = "IEEE Transactions Pattern Analysis and Machine Intelligence",
  volume = "19",
  number = "7",
  pages = "711-720",
  year = "1997"
}

@inproceedings{ charikar97incremental,
  author = "Moses Charikar and Chandra Chekuri and Tomas Feder and
    Rajeev Motwani",
  title = "Incremental Clustering and Dynamic Information Retrieval",
  booktitle = "Proceedings of the 29th Annual ACM Symposium on
    Theory of Computing",
  pages = "626-635",
  year = "1997"
}

@article{ gray98quantization,
  author = "Robert Gray and David Neuhoff",
  title = "Quantization",
  journal = "IEEE Transactions on Information Theory",
  volume = "44",
  number = "6",
  pages = "2325-2383",
  year = "1998"
}

@phdthesis{ tax01thesis,
  author = "David Tax",
  title = "One-Class Classification",
  school = "Tu Delft",
  year = "2001"
}

@article{ wu02face,
  author = "Jianxin Wu and Zhi-Hua Zhou",
  title = "Face Recognition with One Training Image Per Person",
  journal = "Pattern Recognition Letters",
  volume = "49",
  number = "14",
  pages = "1711-1719",
  year = "2002"
}

@article{ zhao03face,
  author = "Wen-Yi Zhao and Rama Chellappa and P. Phillips and
    Azriel Rosenfeld",
  title = "Face Recognition: A Literature Survey",
  journal = "ACM Computing Surveys",
  volume = "35",
  number = "4",
  pages = "399-458",
  year = "2003"
}

@inproceedings{ zhu03semisupervised,
  author = "Xiaojin Zhu and Zoubin Ghahramani and John Lafferty",
  title = "Semi-Supervised Learning Using {Gaussian} Fields and Harmonic
    Functions",
  booktitle = "Proceedings of the 20th International Conference on Machine
    Learning",
  pages = "912-919",
  year = "2003"
}

@inproceedings{ balcan05application,
  author = "Maria-Florina Balcan and Avrim Blum and
    Patrick Pakyan Choi and John Lafferty and Brian Pantano and
    Mugizi Robert Rwebangira and Xiaojin Zhu",
  title = "Person Identification in Webcam Images: An Application of
    Semi-Supervised Learning",
  booktitle = "ICML 2005 Workshop on Learning with Partially Classified
    Training Data",
  year = "2005"
}

@article{ he05face,
  author = "Xiaofei He and Shuicheng Yan and Yuxiao Hu and Partha Niyogi and
    HongJiang Zhang",
  title = "Face Recognition Using {Laplacianfaces}",
  journal = "IEEE Transactions Pattern Analysis and Machine Intelligence",
  volume = "27",
  number = "3",
  pages = "328-340",
  year = "2005"
}

@inproceedings{ beygelzimer06cover,
  author = "Alina Beygelzimer and Sham Kakade and John Langford",
  title = "Cover Trees for Nearest Neighbor",
  booktitle = "Proceedings of the 23rd International Conference on Machine
    Learning",
  pages = "97-104",
  year = "2006"
}

@article{ tan06face,
  author = "Xiaoyang Tan and Songcan Chen and Zhi-Hua Zhou and Fuyan Zhang",
  title = "Face Recognition from a Single Image per Person: A Survey",
  journal = "Pattern Recognition",
  volume = "39",
  number = "9",
  pages = "1725-1745",
  year = "2006"
}

@inproceedings{ goldberg08online,
  author = "Andrew Goldberg and Ming Li and Xiaojin Zhu",
  title = "Online Manifold Regularization: A New Learning Setting and
    Empirical Study",
  booktitle = "Proceeding of European Conference on Machine Learning and
    Principles and Practice of Knowledge Discovery in Databases",
  year = "2008"
}

@inproceedings{ grabner08semisupervised,
  author = "Helmut Grabner and Christian Leistner and Horst Bischof",
  title = "Semi-Supervised On-Line Boosting for Robust Tracking",
  booktitle = "Proceedings of the 10th European Conference on
    Computer Vision",
  pages = "234-247",
  year = "2008"
}

@inproceedings{ sanderson09multiregion,
  author = "Conrad Sanderson and Brian Lovell",
  title = "Multi-Region Probabilistic Histograms for Robust and Scalable Identity Inference",
  booktitle = "Proceedings of the 3rd International Conferences on Advances in Biometrics",
  pages = "199-208",
  year = "2009"
}

@inproceedings{ kveton10semisupervised,
  author = "Branislav Kveton and Michal Valko and Ali Rahimi and
    Ling Huang",
  title = "Semi-Supervised Learning with Max-Margin Graph Cuts",
  booktitle = "Proceedings of the 13th International Conference on
    Artificial Intelligence and Statistics",
  pages = "421-428",
  year = "2010"
}

@inproceedings{ valko10online,
  author = "Michal Valko and Branislav Kveton and Ling Huang and
    Daniel Ting",
  title = "Online Semi-Supervised Learning on Quantized Graphs",
  booktitle = "Proceedings of the 26th Conference on Uncertainty in
    Artificial Intelligence",
  year = "2010"
}

@article{ opencv-library,
  author = "Gary Bradski",
  title = "{The OpenCV Library}",
  journal = "Dr. Dobb's Journal of Software Tools",
  year = "2000"
}

\end{document}